\documentclass[letterpaper, 10 pt, conference]{ieeeconf}
\usepackage{algorithmic}
\usepackage{array}
\usepackage{textcomp}
\usepackage{stfloats}
\usepackage{url}
\usepackage{verbatim}
\usepackage{balance}
\usepackage{diagbox}

\usepackage{amsmath,amssymb,amsfonts}
\usepackage{amsmath}
\usepackage{graphicx}
\usepackage{textcomp}
\usepackage{multirow}
\usepackage{booktabs}
\usepackage{soul,color}

\usepackage{threeparttable}

\usepackage{subfigure} 
\newlength{\whilewidth}
\title{\LARGE \bf
	SoftGPT: Learn Goal-oriented Soft Object Manipulation Skills by \\ Generative Pre-trained Heterogeneous Graph Transformer
} 

\author{Junjia Liu$^{1}$, Zhihao Li$^{1}$, Wanyu Lin$^{2}$, Sylvain Calinon$^{3}$,  \IEEEmembership{Member, IEEE},\\ Kay Chen Tan$^{2}$,  \IEEEmembership{Fellow, IEEE}, and Fei Chen$^{\dagger 1}$, \IEEEmembership{Senior Member, IEEE}% <-this % stops a space
	\thanks{This work was supported in part by the Research Grants Council of the Hong Kong SAR under Grant 24209021, 14222722 and C7100-22GF and in part by CUHK Direct Grant for Research under Grant 4055140.}% <-this % stops a space
	\thanks{$^{1}$Junjia Liu, Zhihao Li and Fei Chen are with the Department of Mechanical and Automation Engineering, T-Stone Robotics Institute, The Chinese University of Hong Kong, Hong Kong SAR (e-mail: {\tt\small jjliu@mae.cuhk.edu.hk, zhihaoli@mae.cuhk.edu.hk, f.chen@ieee.org}).}
	\thanks{$^{2}$Wanyu Lin and Kay Chen Tan are with the Department of Computing, The Hong Kong Polytechnic University, Hong Kong SAR  (e-mail: {\tt\small wanylin@comp.polyu.edu.hk, kctan@polyu.edu.hk}).}
	\thanks{$^{3}$Sylvain Calinon is with the Idiap Research Institute, Martigny, Switzerland  (e-mail: {\tt\small sylvain.calinon@idiap.ch}).}
	\thanks{$^\dagger$Corresponding authors}
}

\begin{document}
		    \IEEEoverridecommandlockouts
%	\overrideIEEEmargins    % Needed to meet printer requirements. 
	\maketitle

	\begin{abstract}
%		Most tasks in the kitchen scene require manipulation of soft objects, which are challenging for existing robotic skill learning techniques due to the complex dynamics and variable shape characteristics of them. In order for the robot to quickly master a variety of soft object manipulation skills through several human demonstration, the robot needs to build a prior knowledge of the representation and dynamics of soft objects. Thus, we propose to construct a pre-trained world model (SoftGPT) that consists of three-dimensional heterogeneous graph representation and generative dynamics model. SoftGPT is trained by massive exploration data and is goal-agnostic. For each downstream tasks, a goal-oriented policy agent is trained to predict the next actions, and SoftGPT generates the consequence of these actions. By doing the two alternately, a thinking process can be built in the robot's mind to provide rollout for policy learning. We demonstrate that with thinking based on the prior knowledge, diverse soft object manipulation skills can be learned efficiently and has the potential to learn directly from several human demonstrations.
	Soft object manipulation tasks in domestic scenes pose a significant challenge for existing robotic skill learning techniques due to their complex dynamics and variable shape characteristics. Since learning new manipulation skills from human demonstration is an effective way for robot applications, developing prior knowledge of the representation and dynamics of soft objects is necessary. In this regard, we propose a pre-trained soft object manipulation skill learning model, namely SoftGPT, that is trained using large amounts of exploration data, consisting of a three-dimensional heterogeneous graph representation and a GPT-based dynamics model. For each downstream task, a goal-oriented policy agent is trained to predict the subsequent actions, and SoftGPT generates the consequences of these actions. Integrating these two approaches establishes a thinking process in the robot's mind that provides rollout for facilitating policy learning. Our results demonstrate that leveraging prior knowledge through this thinking process can efficiently learn various soft object manipulation skills, with the potential for direct learning from human demonstrations.
		
	\end{abstract}
	
%	\begin{IEEEkeywords}
%		Soft object manipulation, pre-trained model, goal-oriented policy
%	\end{IEEEkeywords}
	
	\section{Introduction}

	Soft object manipulation is a fundamental and common skill in human housework that has yet to be fully explored in robotics. Compared with deformable object manipulation that has been well studied \cite{yin2021modeling} \cite{navarro2017fourier} \cite{hu20193}, the shape change of soft objects is not elastic deformation, which makes it more difficult to predict the shape change outcome. Robots that can manipulate soft objects with the same dexterity and efficiency as humans have the potential to improve their usefulness in our daily lives significantly. They can help humans with tasks such as folding laundry, kneading dough, stir-frying, chopping, and prepping vegetables, which are more widespread than deformable object manipulation. However, soft object manipulation currently faces two significant challenges, including the representation of the shape change and the dynamics of soft objects. The shape change of soft objects often has a specific goal, so in order to measure the difference between the current shape and the target shape, it is necessary to propose a compact representation that can cope with the shape and thickness variations of soft objects. Furthermore, although recent works have shown the potential of generating a world model with environment dynamics to efficiently learn robot manipulation skills online \cite{wu2022daydreamer}, these models still primarily focus on rigid objects. The dynamics of soft objects that produce shape change are more complex, and learning directly in real-world scenarios can be time-consuming due to the reset. 
		\begin{figure}[t]
		\centering
		\includegraphics[width=0.85\linewidth]{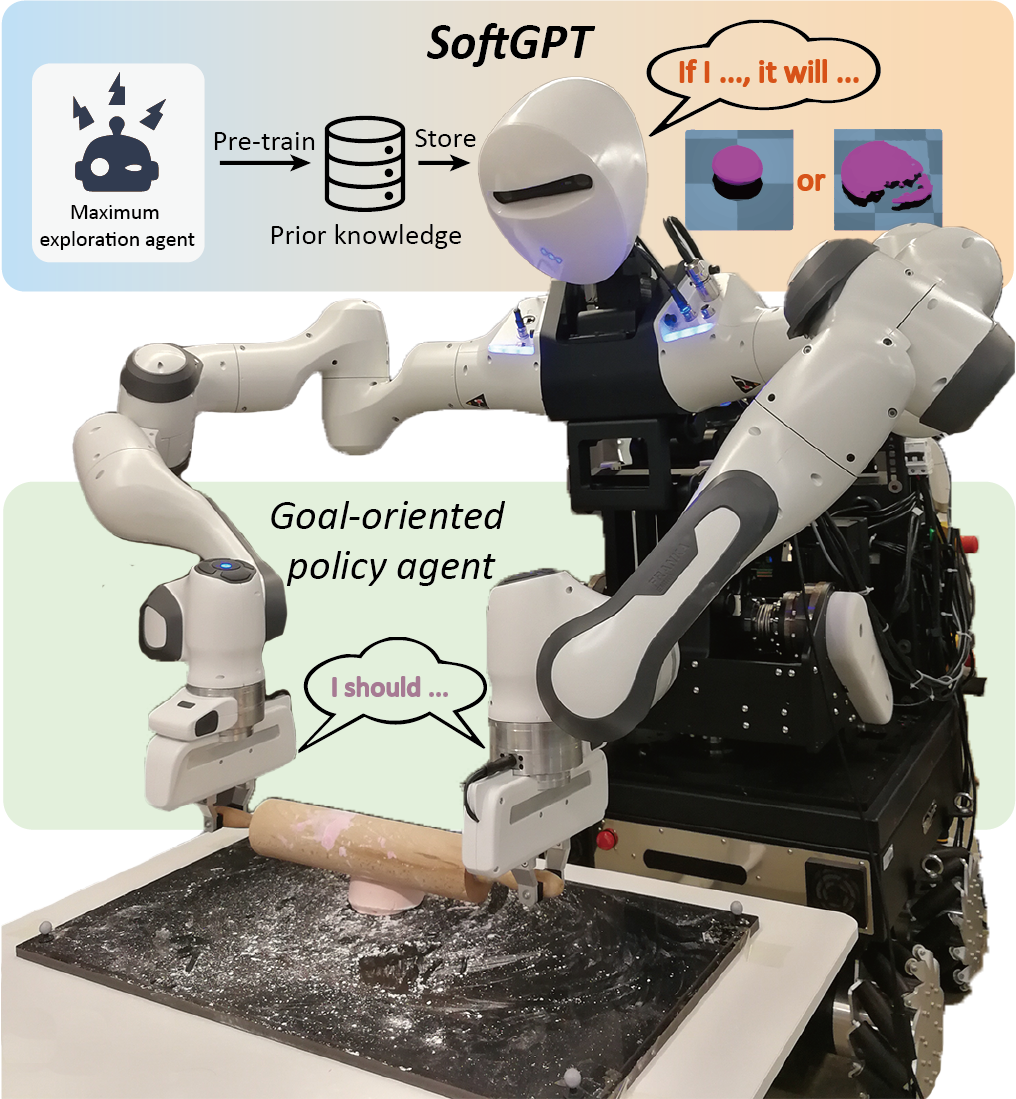}
		\caption{The proposed pipeline of soft object manipulation contains two parts: a pre-trained SoftGPT model for providing prior knowledge about soft object manipulation and a goal-oriented policy for each specific task.}
		\label{cover}
		\vspace{-1em}
	\end{figure}
    
    In the domestic scenario, the ability to learn these complex skills directly from a few human demonstrations will be the key to the development of domestic service robots \cite{liu2022robot}, which suggests that robots should have a fundamental judgment on the manipulation \cite{liu2020efficient} \cite{liu2023revolt} in addition to the demonstration data. Drawing inspiration from the hot topics in natural language processing, we propose to develop a Generative Pre-trained Transformer (GPT) \cite{radford2019language} based model to establish a prior knowledge about the representation and dynamics of soft objects, supporting the efficient skill learning of soft object manipulation. Compared with the recent discussion about using ChatGPT as high-level function commands \cite{vemprala2023chatgpt}, we tend to find ways to integrate GPT technologies more closely with robotics. Thus, in this paper, we adopt a GPT-based model named SoftGPT to learn the relationship between robot manipulation and soft object shape change, similar as how GPT works when answering questions. Generally speaking, it serves as a pre-trained world model in typical robotics research. As shown in Fig. \ref{cover}, the SoftGPT with prior knowledge can estimate the manipulation consequence in advance and rehearse a rollout in robots' mind, and supports the skill learning of specific goal-oriented policies. Such a pre-trained model can significantly reduce the computational and time costs of training models from scratch. 
    
	Despite the help of differential simulators, learning soft object manipulation from scratch directly using reinforcement learning (RL) has proven difficult. PlasticineLab, a differentiable elastic and plastic deformation benchmark utilizing the DiffTaichi system \cite{hu2019difftaichi}, is an effective tool for evaluating the efficiency of RL-based approaches in solving shaping tasks \cite{huang2021plasticinelab}. The baseline results it provided demonstrate that RL-based methods cannot solve most shaping tasks in this context efficiently. Based on this work, RoboCraft is proposed to model a particle-based dynamics for elasto-plastic objects using Graph Neural Network (GNN) \cite{RoboCraft}. We use the differential simulator to generate massive interactive data to pre-train the SoftGPT model and obtain the dynamics based on the graph representation of shape changes.

		\begin{figure*}[t]
	\centering
	\includegraphics[width=0.95\linewidth]{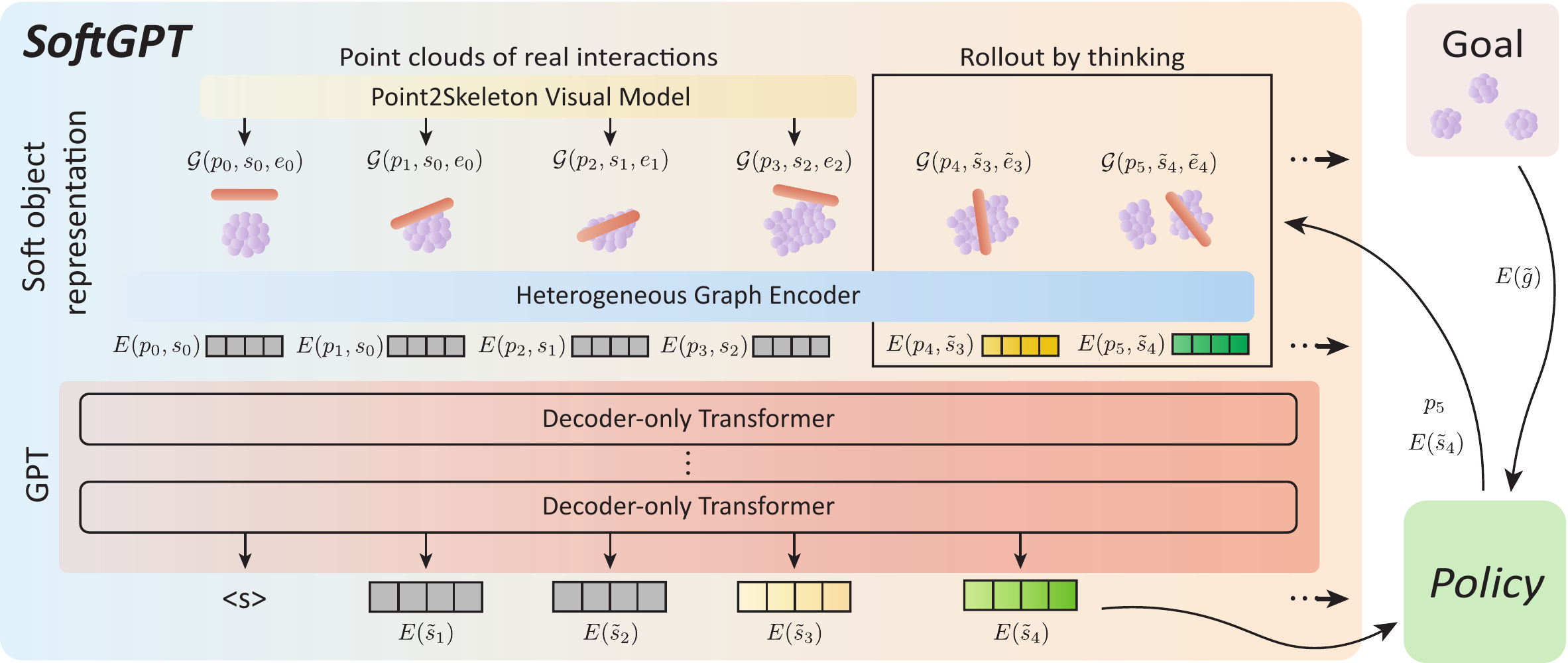}
	\caption{SoftGPT is a pre-trained world model that learns the representation of soft objects and to predict the dynamics of environment. The model processes point clouds obtained from real interactions and represents them as three-dimensional compact graphs using the Point2skeleton Visual Model. With the next robot pose provided by a goal-oriented policy, the model predicts the next state using several Decoder-only Transformer blocks. In this regard, the purple balls represent skeleton points, and the orange bars represent operation nodes, which illustrate a policy learning process for a cutting task with the aid of SoftGPT.}
	\label{softgpt}
	\vspace{-1em}
\end{figure*}

	The main contributions of this paper can be summarized as follows:
	\begin{itemize}
		\item \textbf{SoftGPT model:} We propose a generative pre-trained Transformer-based model (SoftGPT) trained by massive interaction data for compact representation and dynamics modeling of soft objects.
		\item \textbf{Goal-oriented policy agent:} We provide a goal-oriented policy for each downstream tasks. By alternately performing the policy  and SoftGPT, a thinking process is built in the robot's mind to provide rollout for policy learning.
		\item \textbf{Soft object manipulation:} We demonstrate that the pre-trained SoftGPT model can accelerate the learning of downstream policies, makes it efficient to learn task-specific soft object manipulations.
		
	 \end{itemize}

	\section{Generative Pre-trained Heterogeneous Graph Transformer and Goal-oriented Policy Agent}
	The present study proposes a pipeline for robot manipulation of soft objects (Fig. \ref{cover}), comprising two main stages: pre-training and thinking while learning. The pre-training stage involves training a SoftGPT model using a dataset generated by a maximum exploration agent. This SoftGPT model serves as a world model of soft objects, including the prior knowledge about manipulation consequences of them. The thinking stage employs a model-based policy learning mechanism that alternatively predicts the next action and state using both a goal-oriented policy agent and the SoftGPT model. The key idea behind this approach is to provide the robot with a fundamental understanding of soft objects through extensive contact with the soft object environment. This approach allows the robot to leverage prior knowledge when learning specific manipulation skills and avoid exploring meaningless actions. In summary, the proposed pipeline seeks to enhance the robot's cognitive capabilities in manipulating soft objects and enable it to learn new skills more efficiently.
	
%	The schematic diagram of the whole pipeline is illustrated in Fig. \ref{pipeline_simple}, which consists of two parts: a pre-training process and a thinking process. During the pre-training process, the SoftGPT is trained with a dataset that generated by a maximum exploration agent, serves as a world model of soft objects. Then, the thinking process refers to a model-based policy learning mechanism that predicts the next action and the next state alternatively by both the goal-oriented policy agent and the goal-agnostic SoftGPT model. The key concept of this idea is to give the robot a basic cognition of soft objects through full contact with the soft object environment, so that it can think twice before the real interaction based on prior knowledge when learning specific manipulation skills and avoid the exploration of meaningless actions. 
	
	In this section, we first elaborate the structure of SoftGPT model, including dynamic modeling by GPT and the corresponding graph-based shape representation of soft objects  (Sec. \ref{IIA}). Then we show how to construct a goal-oriented policy agent in the latent space of the representation, and describe how it alternates with SoftGPT during policy learning to form the robot's thinking process (Sec. \ref{IIB}).

	\subsection{Learn contact dynamics model from massive interaction}\label{IIA}
%	The core module of SoftGPT is a comprehensive dynamics modeling model, consists of an encoder based on a heterogeneous graph neural network and stacked decoder-only Transformers. The details of SoftGPT are illustrated in Fig. \ref{softgpt}. 
%	
%	SoftGPT is pre-trained in an unsupervised manner like GPT-2 \cite{radford2019language}, namely the model is trained to predict the next object state in a sequence based on the previous states and actions. However, unlike the typical sequence-to-sequence tasks in natural language processing, there have two heterogeneous roles (manipulator and object) in the robot manipulation task. We need to do some additional processing on the training set so that it can meet the goal of predicting future states, that is, construct a shifted dataset based on the exploration graph dataset after representing by the Point2Skeleton visual model. Specifically, the manipulator nodes are shifted right to form a graph like $\mathcal{G}(p_{t+1}, s_t, e_t)$. Since the edges between the manipulator nodes and object nodes are fully-connected and constant, the time annotation of edges follow the change of object nodes. Take a sequence of graph samples in this shifted dataset as input, a heterogeneous graph encoder is set to convert graphs to latent embedding vectors $E(p_{t+1}, s_t)$. Then this sequence of latent vectors are regarded as past experience and inputted to a stacked decoder-only Transformer to predict the embedding of next states $E(\tilde s _{t+1})$. 

The core module of SoftGPT is a GPT-based dynamics modeling model, which consists of a heterogeneous graph convolutional neural network-based \cite{kipf2016semi} encoder and stacked decoder-only Transformers. Fig. \ref{softgpt} provides an illustration of its components. SoftGPT is pre-trained in an unsupervised manner, similar to GPT-2 \cite{radford2019language}. It is trained to predict the next state of the object in a sequence based on the previous states and actions. However, unlike typical sequence-to-sequence tasks in natural language processing, robot manipulation tasks involve two heterogeneous roles: the manipulator and the object. To accommodate this, an additional processing should be first performed on the training set, that is, to construct a shifted dataset based on the exploration graph dataset represented by the visual model. Specifically, manipulator nodes should be shifted right to form a graph like $\mathcal{G}(p_{t+1}, s_t, e_t)$. Since the edges between the manipulator nodes and object nodes are fully-connected and constant, the time annotation of edges follows the change of object nodes. Given a sequence of graph samples in this shifted dataset as input, a heterogeneous graph encoder converts the graphs to latent embedding vectors $E(p_{t+1}, s_t)$. These latent vectors are then considered as past experience and inputted to a stacked decoder-only Transformer to predict the embedding of the next state, $E(\tilde s_{t+1})$. Soft objects deform as a result of the continuous movement of a robot, and the state feature of the manipulator node only captures the current pose. Hence, it is reasonable to replace the LSTM-based Recurrent State-Space Model (RSSM) \cite{hafner2019learning} in some world model works \cite{wu2022daydreamer}\cite{hafner2019dream} with Transformer \cite{vaswani2017attention} that can handle longer sequences of historical manipulations.

\begin{figure}[t]
 	\centering
 	\includegraphics[width=1\linewidth]{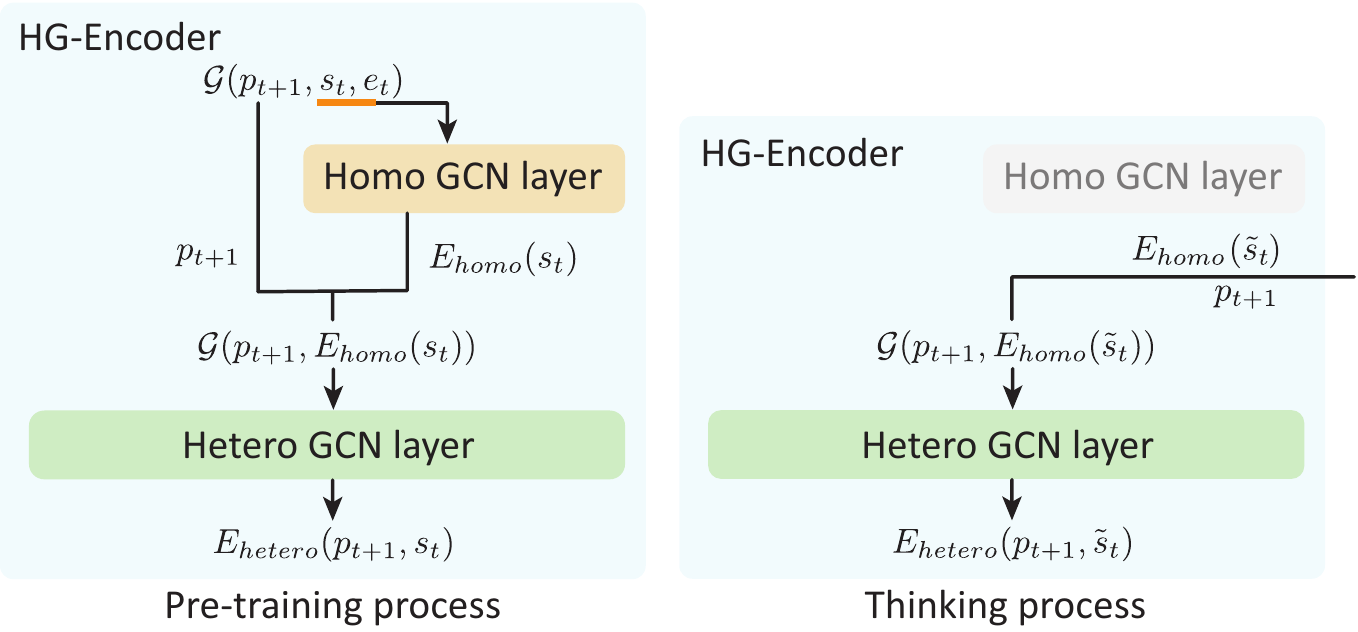}
 	\caption{Heterogeneous Graph Encoder in pre-training and thinking process. Left: The object subgraph in a real interaction is encoded first and a new graph with embedding object features is built to obtain the final graph embedding. Right: Embedding object features and the action are provided by the SoftGPT and policy, the homogeneous layer is skipped.
 	}
 	\label{HGEncoder}
 	\vspace{-1em}
 \end{figure}

%The decoder-only Transformer model largely follows the details of the OpenAI GPT-2 model \cite{radford2019language} with a few modifications. Since it is time-consuming and meaningless to reconstruct an accurate heterogeneous graph data with continuous variable features from hidden variables, we learn dynamics directly within the hidden space of the graph encoder. At the same time, in order to make SoftGPT a task-independent pre-trained model, it does not generate the same format as inputs, which is heterogeneous graphs, but only representations of object subgraphs. This also makes the heterogeneous graph neural network encoder need to have some extra tricks to handle this situation. In this encoder, first there is a layer of homogeneous graph neural network only for object subgraphs, and then a heterogeneous graph neural network for the entire scene graph to learn the relationship between manipulator nodes and object subgraphs. Thus, the next embedding prediction of the object's state is fed between these two graph neural networks to combine with the next action, whether derived from data or generated from the policy agent.
 Given that reconstructing accurate heterogeneous graph data with continuous features from hidden vectors is both time-consuming and meaningless, we opted to learn dynamics directly within the hidden space of the graph encoder. To ensure that SoftGPT remains a goal-agnostic pre-trained model, it only generates representations of object subgraphs rather than the same format as inputs, which are heterogeneous graphs. As a result, the heterogeneous graph neural network encoder needs some additional techniques to handle this situation. Specifically, the GNN-based encoder consists of a layer of homogeneous graph neural network solely for object subgraphs, followed by a heterogeneous graph neural network for the entire scene graph, to learn the relationship between manipulator nodes and object subgraphs. Additionally, the next embedding prediction of the object's state is fed between these two graph neural networks and combined with the next action, whether derived from data or generated from the policy agent. The details are illustrated in Fig. \ref{HGEncoder}.

% 	\subsection{Three-dimensional graph representation of soft objects}\label{IIA}

 %	The front module of SoftGPT is a graph representation model, serves as a visual module to obtain compact representations of soft object shapes. Since the deformation of soft objects is three-dimensional, the most straightforward way to describe it is using point clouds. However, most of the dense point cloud information is meaningless and will increase the size of the network and the difficulty of training. Therefore, we should extract representative skeleton points from the point cloud and represent the shape characteristics of the object with the position of skeleton points and the radius of the sphere centered on them. We draw on the ideas of the Point2Skeleton work \cite{lin2021point2skeleton} and use Medial Axis Transformation (MAT) \cite{blum1967transformation} technique to extract skeleton points.
 
 To support the efficient process of Heterogeneous Graph Encoder, a visual model is required to obtain compact and effective representations of soft object shapes. We use Point2Skeleton model \cite{lin2021point2skeleton} to extract three-dimensional graph with a small number of nodes from dense point clouds. The node features are the positions and radii of the extracted skeletal points. This graph is used as a subgraph denoted as $\mathcal{G}(s_t), t\in\mathbb{R}^+$, to represent the shape characteristics of the soft object. When combined with the manipulator node, which includes the feature of its pose, denoted as $\mathcal{G}(p_t)$, it forms a heterogeneous graph $\mathcal{G}(p_t, s_t, e_t)$ that depicts the scene. Here, $e_t$ refers to the edges connecting the nodes in the graph. 
 
 %	With the Point2Skeleton visual model, point clouds can be extracted as a graphs with node features of skeletal point positions and their radii. This graph is used as a subgraph $\mathcal{G}(s_t), t\in\mathbb{R}^+$ for representing the shape characteristics of the soft object, and together with the manipulator node with the feature of its pose $\mathcal{G}(p_t)$, it forms a heterogeneous graph $\mathcal{G}(p_t, s_t, e_t)$ depicting the scene, where $e_t$ refers to the edges.

 %	This approach allows for the representation of complex scenes using heterogeneous graphs, where each node can represent a different entity in the environment, and the edges capture the relationships between them.

\subsection{Learn goal-oriented policy with goal-agnostic SoftGPT}\label{IIB}

In the policy learning process, SoftGPT functions as a pre-trained world model with updates that occur infrequently. As depicted in Fig. \ref{softgpt}, SoftGPT takes the embedding sequence from real interactions as input and predicts the consequence $s_{t+1}$ of pose $p_{t+1}$ at state $s_t$. The policy agent has the ability to generate actions based on either real states or predicted states in rollout. The data flow during the policy learning process is represented by the curves in Fig. \ref{softgpt}.

The policy agent operates on states in hidden space and includes encoded goal states, predicting the next pose of the manipulator. Specifically, the policy agent uses the actor-critic framework \cite{haarnoja2018soft} with double Q-network \cite{van2016deep}, and the training of these two networks employs data from both real interactions and SoftGPT rollout by the thinking process. The modified loss functions are defined as follows:
 \begin{equation}
	 	\begin{aligned}
		 	 &\text{\textit{Policy}}\\
		 		& \mathcal{L}_\pi =  -\mathbb{E}\Big[\min _{i=1,2} \widetilde Q_{\phi_i}\left(\varepsilon_t, \tilde{a}_\theta(\varepsilon_t)\right) + \eta(\varepsilon_t|E(g)) +\alpha \mathcal{H}\left(\cdot \mid \varepsilon_t\right)\Big]\\
		 		&\text{\textit{Q functions}}\\
		 		 	&\mathcal{L}_Q = \mathbb{E} \Big[ Q(\varepsilon_t, a)- \Big(r + \eta(\varepsilon_{t+1}|E(g)) + \alpha \mathcal{H}\left(\cdot \mid \varepsilon_{t+1})\right) 
		 		 	\\&\quad\qquad \gamma(1-d)\Big(\min _{i=1,2} \widetilde Q_{\phi_{targ,i}}\left(\varepsilon_{t+1}, \tilde{a}_\theta(\varepsilon_{t+1})\right)\Big)\Big) \Big]^2 
		 	\end{aligned}
	 \end{equation}

\begin{figure*}[t]
	\centering
	\includegraphics[width=0.9\linewidth]{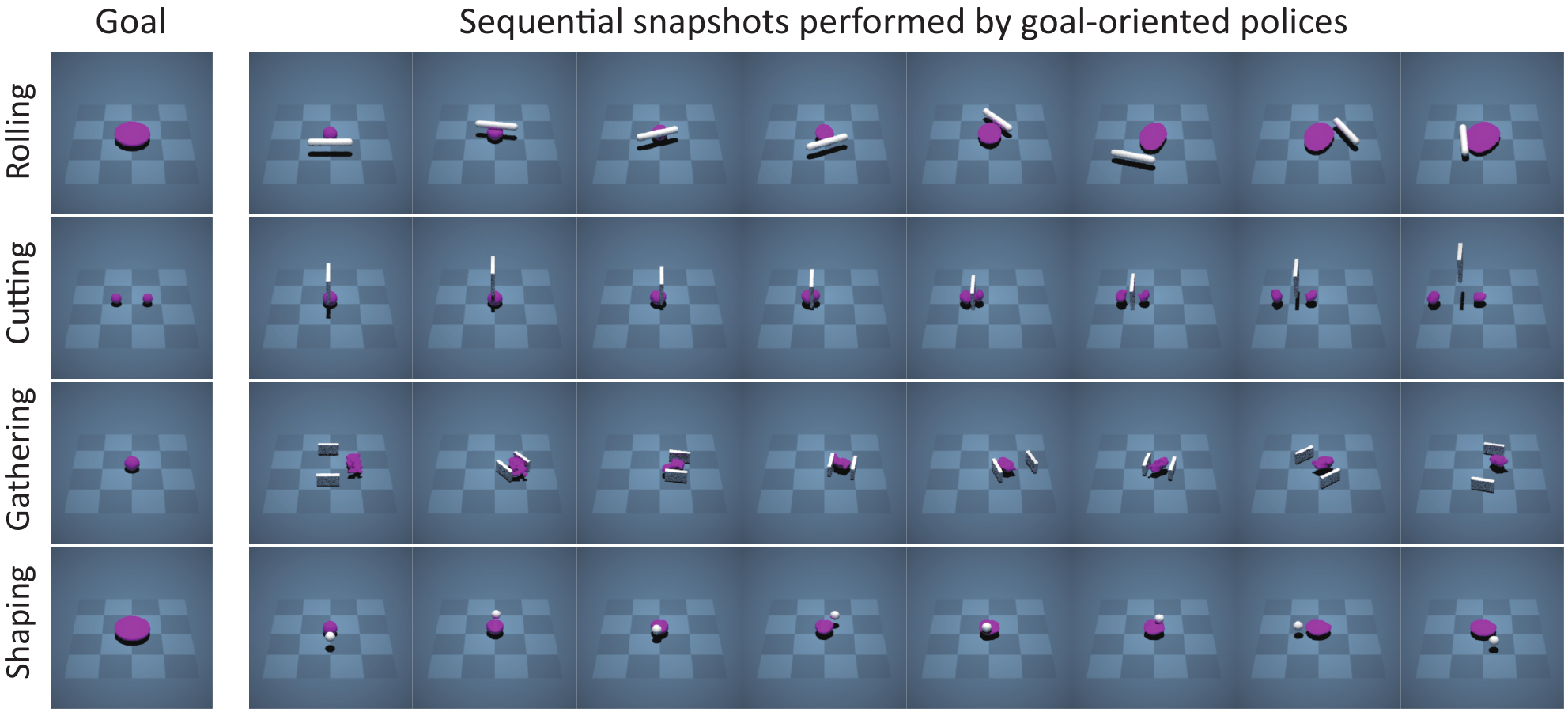}
	\caption{Sequential snapshots performed by goal-oriented policy for each specific task.}
	\label{Exp}
	 	\vspace{-1em}
\end{figure*}

Here, $\varepsilon_t = E(s_t)$, $d$ represents the done signal, $r$ refers to the real reward, and $\mathcal{H}$ refers to the entropy of states. $\eta(\cdot|E(g))$ is a learned goal-oriented reward function in the latent space. $\widetilde Q_{\phi_{i}}\left(\varepsilon_{t+1}, \tilde{a}\theta(\varepsilon{t+1})\right)$ refers to the modified Q evaluation that incorporates the thinking process:
 \begin{equation}
		\begin{aligned}
		 (1- \lambda)Q_{\phi_i}\left(\varepsilon_t, \tilde{a}_\theta(\varepsilon_t)\right) + \lambda Q_{\phi_i}^\lambda\left(\tilde\varepsilon_{t+1}\right)
		 	\end{aligned}
 \end{equation}

The above equation is averaged over all $N\in[t+1, t+H]$ with $\lambda$ discount, where $H$ is the thinking horizon, and $\tilde\varepsilon_N$ refers to one of the rollout states generated from the thinking process. In this process, the policy alternates with the SoftGPT model, following the loop:
 \begin{equation}
	\begin{aligned}
		\tilde\varepsilon_{t+1} = \text{SoftGPT}(\varepsilon_{t}, p_{t+1}) \rightarrow p_{t+2} = \pi(\tilde\varepsilon_{t+1})\rightarrow \dots \rightarrow \tilde\varepsilon_{t+H}
	\end{aligned}
\end{equation}

The incorporation of additional rollout from the thinking process enables the policy to consider long-horizon results, thereby improving the efficiency of policy learning through increased interaction in the robot's cognitive process.

	\section{Experiments}\label{III}

	\subsection{Setup}\label{IIIA}
	\textbf{Tasks:} In order to demonstrate that the proposed pre-trained model can promote efficient learning with downstream tasks, we design four tasks with different tools and initial states:
	\begin{itemize}
		\item \textit{Rolling} Train an agent to roll and spread a ball-shaped dough into a specific cake state using a rolling pin.
		\item \textit{Cutting} Train an agent to separate a ball-shaped dough into several specific parts using a flat knife.
		\item \textit{Gathering} Train an agent to use two flat-shaped tools to aggregate multiple scattered dough into a specific shape.
		\item \textit{Shaping} Train an agent to use a sphere-shaped rolling ball to shape the dough into a specific cake state.
	\end{itemize}

	The action dimension of rolling pin, knife and rolling ball are 3, while the action dimension of dual flats is 7.

%	\begin{table}
%	\centering
%	\caption{The composition of exploration dataset}
%	\label{table1}
%	\begin{threeparttable}
%		\begin{tabular}{p{1.3cm}p{1.3cm}p{1.3cm}p{1.3cm}p{1.3cm}}
%			\toprule[1pt]
%			\diagbox[innerwidth=1.45cm]{Shapes}{Tools} 			& Rolling Pin 	& Knife & Dual Flats  & Rolling Ball   \\
%			\midrule
%			Ball 			& 1489 		& 1363 	 & - 	  & 1693  \\
%			Two balls				& - 	& - & 1061 & - \\
%			Cuboid				& 1328		& 1036 & - & 1124\\
%			Random	& - 	& - & 1169 & -\\
%			\bottomrule[1pt]
%		\end{tabular}
%		\begin{tablenotes}
%			\footnotesize
%			\item $^\dagger$ The numbers refer to the amount of interactive data with specific tools and different initial shapes.
%		\end{tablenotes}
%	\end{threeparttable}
%\end{table}

	\textbf{Simulation:} The simulation environment is built on top of PlasticineLab \cite{huang2021plasticinelab}, with a particle-based dough and tools listed in the tasks. We take the position of the particles that make up the dough from the simulator and pass their surface particles into the network as simulated point cloud information. 
	
%	\textbf{Evaluation metric} Since the full state of particles is available in the simulator, we compare the Signed Distance Field (SDF), density, and Intersection over Union (IoU) between the current state and the target state.

	\textbf{Evaluation metric:} In the simulator, the complete state of particles is accessible, allowing us to evaluate the current state of the system in comparison to the target state using various metrics. Precisely, we assess the Signed Distance Field (SDF), density, and Intersection over Union (IoU) between the two states, forming the reward function together.
	
	\subsection{Exploration dataset generation}\label{IIIB}
%	There are two approaches to generate exploration dataset for the pre-trained model. We can ask the demonstrator to wear an Xsens motion capture suit to collect human motion information, and set up a multi-view camera to collect point clouds of soft objects. However, working directly with real-world data presents several engineering issues that need to be addressed, such as occlusion between objects, the quality of point clouds, and so on. Considering that the contribution of this work lies in exploring the role of pre-trained models on downstream soft object manipulation skill learning, we chose to use a simple reinforcement learning based explore agent in differential simulators to generate interactive data. In order for it to fully interact with the soft object, we regard the initial state of the soft object as the target and regard the loss as reward to encourage the agent to shape the object as much as possible from the initial state. During the interaction, the point clouds of the object and the pose of the manipulator are recorded in sequence for pre-training the graph representation model and the dynamics GPT-based model.  

	\begin{figure*}[t]
	\centering
	\includegraphics[width=0.9\linewidth]{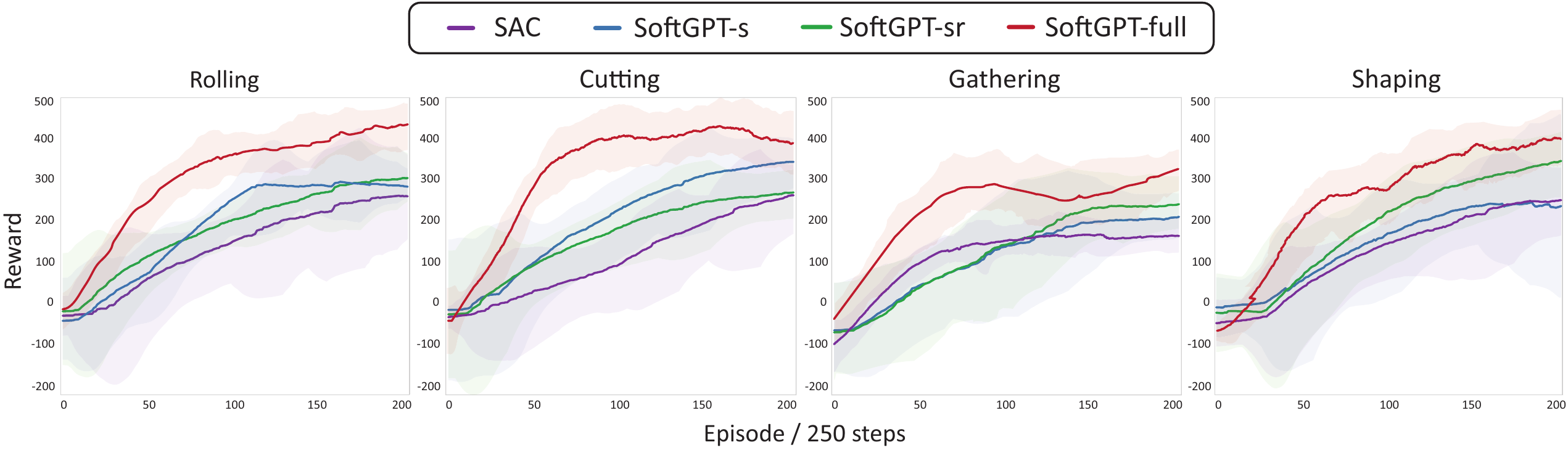}
	\caption{The comparison of downstream task learning efficiency between using pre-trained SoftGPT and baseline methods. Baseline methods contain original Soft Actor-Critic policy (SAC), Soft Actor-Critic with SoftGPT from scratch (SoftGPT-s) and Soft Actor-Critic with SoftGPT from scratch and goal-oriented reward function $\eta$ (SoftGPT-sr). All methods share the same visual model as state representation and Heterogeneous Graph Encoder.}
	\label{learning}
		 	\vspace{-1em}
\end{figure*}

	\begin{table}
	\centering
	\caption{The composition of exploration dataset}
	\label{table1}
	\begin{threeparttable}
		\begin{tabular}{p{1.3cm}|p{1.3cm}p{1.3cm}p{1.3cm}p{1.4cm}}
			\toprule[1pt]
			{Shapes} 			& Rolling Pin 	& Knife & Dual Flats  & Rolling Ball   \\
			\midrule
			Ball 			& 11398 	& 8973 	& - 	  	& 8083  \\
			Two balls		& - 		& - 	& 2980 		& - \\
			Cuboid			& 7937		& 5072 	& - 		& 8923\\
			Random			& - 		& - 	& 4000 		& -\\
			\bottomrule[1pt]
		\end{tabular}
		\begin{tablenotes}
			\footnotesize
			\item $^\dagger$ The numbers refer to the amount of interactive data with specific tools and different initial shapes.
		\end{tablenotes}
	\end{threeparttable}
		 	\vspace{-1em}
\end{table}

The generation of an exploration dataset for pre-trained models can be approached through two methods. One involves collecting human motion information using a motion capture system like Xsens suit and capturing point clouds of soft objects using multi-view cameras. However, this approach poses several engineering challenges, such as occlusion between objects and ensuring the quality of point clouds. Given that this work focuses on exploring the impact of pre-trained models on downstream soft object manipulation skill learning, we opted to use a RL-based exploration agent to generate interactive data in differential simulators. To facilitate full interaction with the soft object, we consider the initial state of the object as the target and frame the loss as a reward to incentivize the agent to shape the object from the initial state as much as possible. The initial state was randomly generated, incorporating both aggregated and scattered patterns to prepare for subsequent gatherings and cutting tasks. During the interaction, we record the sequence of point clouds of the object and the pose of the manipulator for pre-training both the graph representation model and the dynamics model. We screened out data that came into contact with objects, and the composition of the exploration dataset is listed in Table \ref{table1}. 

%	
%	\begin{table}
%		\centering
%		\caption{\alert{The composition of exploration dataset}} %添加标题 设置标签
%		\label{table1}
%		\begin{threeparttable}
%			\begin{tabular}{p{1.3cm}p{1.3cm}p{1.3cm}p{1.3cm}p{1.3cm}}
%				\toprule[1pt]
%				Tools 			& Rolling pin 	& Flat & Dual flats  & Sphere   \\
%				\midrule
%				Ball 			& - 		& - 	 & - 	  & -  \\
%				Cuboid				& 63.7		& 0.0733 & 1.5307 & 327.9\\
%				Two balls				& 509.3 	& 0.3883 & 0.9056 & 284.5 \\
%				Three balls 				& 629.5 	& 0.4203 & 0.6422 & 271.9 \\
%				Random	& \textbf{756.1} 	& \textbf{0.5573} & \textbf{0.6223} & \textbf{266.4}\\
%				\bottomrule[1pt]
%			\end{tabular}
%			\begin{tablenotes}
%				\footnotesize
%				\item $^\dagger$ The numbers refer to the amount of interactive data with specific tools and different initial states.
%			\end{tablenotes}
%		\end{threeparttable}
%	\end{table}
	
	\subsection{Implementation details}\label{IIIC}
	\textbf{Maximum exploration agent:} This agent is implemented directly based on the typical soft actor-critic algorithm shared by Rofunc \cite{liu2023rofunc}.
	
	\textbf{Point2Skeleton visual model:} We draw on the ideas of the Point2Skeleton approach \cite{lin2021point2skeleton} and utilize the Medial Axis Transformation (MAT) technique \cite{blum1967transformation} to extract the skeleton points. The representation model takes the point cloud of soft objects as input and learns a geometric transformation to extract skeletal points and their radii. To achieve this, the PointNet++ approach \cite{qi2017pointnet++} is utilized as the encoder to obtain the sampled input points and their contextual features. The convex combination of input points is then used to generate the skeletal points. To ensure the reliability of the resulting representations, a set of reliable links is generated based on topology prior, where a node has a link to its closest node rather than its \textit{k}-farthest nodes. As we do not require mesh information, the further link prediction model in the original work is omitted for the sake of efficiency. For additional details on the Point2Skeleton approach, please refer to \cite{lin2021point2skeleton}. The shell of particles is extracted by alpha shape algorithm \cite{edelsbrunner1983shape}, and used as point cloud data input of the visual model. The model is set to generate 30 skeleton points. 
	
	\textbf{Heterogeneous graph encoder:} The encoder is composed of one homogeneous graph convolutional network layer and one heterogeneous graph convolutional network layer, both of which have a hidden state with a dimension of 32.
	
	\textbf{Stacked Transformer:} The stacked decoder-only Transformer is based on the bare GPT-2 model shared by \cite{wolf-etal-2020-transformers}, which outputs raw hidden-states without any specific head on top. It has 12 hidden layers with a hidden dimension of 32, and 4 attention heads for each attention layer.
	
	\textbf{Goal-oriented policy:} The policy is implemented based on SAC with several modifications, with a hidden dimension of 256 and learning rates of 0.0003. In order to handle the input of graph data, both actor and critic have a preprocessing model composed of a heterogeneous graph neural network, which encodes graph data into latent space vectors. The output of policy is the next pose of tools, which means a trajectory planning between poses is necessary. We adopt Linear Quadratic Tracking (LQT) shared by \cite{liu2023rofunc} to plan a 50-point trajectory between each two poses, and it will be smoothed according to the previous series of points.

	The actor $\pi$, critic $Q_{\phi_i}$, reward model $\eta$ and SoftGPT model are trained using the Adam optimizer. Among them, the actor, critic and latent reward model are updated per 250 steps. SoftGPT model is updated per 500 steps.

	\subsection{Goal-oriented policies for downstream tasks}\label{IIID}
	For each specific task, a goal-oriented policy is trained with the help of the SoftGPT model. We use these learned policies to evaluate in the test environment of the same task, and the results are shown in Fig. \ref{Exp}. It can be observed that all four policy agents manipulate the dough towards their target shapes. The effect of rolling task is the best, which is largely related to the difficulty of the task. On the contrary, the effect of the other three tasks is poor, especially the gathering task. This might be due to the fact that the action dimension of dual flats is higher, which makes the manipulation more difficult. Moreover, the exploration agent is more inclined to break the dough when collecting interaction data, which is not conducive to learning the aggregation task.

%	The pre-trained SoftGPT model allows for efficient learning and enables the robot to adapt to novel situations quickly. The results demonstrate the effectiveness of our proposed method in learning diverse soft object manipulation skills through several human demonstrations.
%	

	\subsection{Learning efficiency analysis of downstream tasks}\label{IIIE}
	\textbf{Baselines:} We conducted an ablation study by comparing our entire proposed method to three baselines.
	
	\begin{itemize}
		\item \textit{Soft Actor-Critic (SAC)}: a SAC-based policy.
		\item \textit{Soft Actor-Critic with SoftGPT from scratch (SoftGPT-s)}: a SAC-based policy with a learned reward function.
		\item \textit{Goal-oriented Soft Actor-Critic with SoftGPT from scratch (SoftGPT-sr)}: a SAC-based policy with a learned reward function and a thinking process supported by a non-pretrained SoftGPT that learns form scratch alongside the policy.
	\end{itemize}
	
	All baselines were trained using the Point2Skeleton visual model and took the heterogeneous graph representation as input. Additionally, all SAC-based policies have graph preprocessing models in their actor and critic models. 

	The comparative performance of the proposed method and its baselines is presented in Fig. \ref{learning}. Notably, the entire proposed method leveraging a pre-trained SoftGPT model (\textit{SoftGPT-full}) exhibits superior efficiency in learning downstream tasks. Its stable performance basically appears around 50-100 episodes, which is two times faster than others. Compared to the baselines, the proposed method achieves higher rewards in all the tested tasks. In particular, the proposed method shows significant improvement in the task of rolling and cutting dough toward a target state, indicating its potential in learning manipulation skills for soft objects. These results demonstrate the effectiveness of using a pre-trained world model for soft object manipulation tasks and suggest that the proposed method has the potential to support skill learning from a few human demonstrations.

\section{Discussion}
	This work still has some limitations. First, this paper focuses on validating the role of a GPT-based pre-trained model for the robot skill learning of downstream tasks in soft object manipulation. We will introduce data augmentation method \cite{liu2023birp} to the dataset so that it can support learning new manipulation skills directly from human demonstrations. And we will verify it on real robots by using meta-learning in the graph representation \cite{Wang_Cao_Lin_Jiang_Tan_2023}, finding the sim2real effect of SoftGPT that is trained with massive simulated interaction data. Second, we can extend the feature of manipulator nodes to add haptics or force information so that the robot can manipulate soft objects with different physical properties based on the relationship between feedback and deformation in the pre-trained model. Finally, this work only draws on the large-scale model and generative concepts of GPT-2. We plan to introduce concepts such as human prompt and RLHF in the future so that humans can guide the manipulation of robots in real-time.
	
	\section{Conclusion}
	This paper focuses on how robots understand, represent, and manipulate soft objects. To support efficient skill learning of soft object manipulations, we propose that robots should have prior knowledge of soft object dynamics and representation. Thus, we introduce the idea of GPT-2 and design a generative pre-trained model with a three-dimensional graph representation for soft objects. With the help of this pre-trained model, the policy of each downstream task can be obtained easily and efficiently, making it possible for humans to teach robots similar skills through several demonstrations.

	\bibliographystyle{ieeetr}
	\bibliography{ref_hg}
\end{document}